\documentclass[runningheads]{llncs}
\usepackage[T1]{fontenc}    
\usepackage{hyperref}       
\usepackage{url}            
\usepackage{booktabs}       
\usepackage{amsfonts}       
\usepackage{nicefrac}       
\usepackage{microtype}      
\usepackage{xcolor}         
\usepackage{times}

\usepackage{soul}
\usepackage{url}
\usepackage[symbol]{footmisc}
\usepackage[misc]{ifsym}

\usepackage[small]{caption}
\usepackage{graphicx}
\usepackage{booktabs}
\usepackage{amssymb}
\usepackage{amsmath}
\usepackage[ruled]{algorithm2e}
\usepackage{algorithmic}
\usepackage{graphicx}
\usepackage{makecell}
\newcommand\Tstrut{\rule{0pt}{2.5ex}}       
\newcommand\Bstrut{\rule[-1.1ex]{0pt}{0pt}} 
\newcommand{\TBstrut}{\Tstrut\Bstrut} 
\usepackage{lipsum}
\usepackage[page]{appendix} 

\usepackage{xcolor}
\definecolor{mycolor}{RGB}{0,20,115}
\hypersetup{
	colorlinks=true,
	linkcolor=mycolor,
	filecolor=magenta,
	urlcolor=mycolor,
	citecolor=mycolor
}

\begin{document}
\title{Imitation learning with Sinkhorn Distances}
\toctitle{Imitation learning with Sinkhorn Distances}

%
%
\author{Georgios Papagiannis\textsuperscript{1,}\thanks{\noindent Work done as a student at the University of Surrey}{\Letter} and Yunpeng Li\textsuperscript{2}}

\authorrunning{G. Papagiannis and Y. Li}

\tocauthor{Georgios Papagiannis\textsuperscript{1,}\thanks{\noindent Work done as a student at the University of Surrey}{\Letter} and Yunpeng Li\textsuperscript{2}}

%
\institute{\textsuperscript{1}Imperial College London\\ \textsuperscript{2}University of Surrey\\\small{\tt g.papagiannis21@imperial.ac.uk, yunpeng.li@surrey.ac.uk}}

\maketitle

\begin{abstract}
 Imitation learning algorithms have been interpreted as variants of divergence minimization problems. The ability to compare occupancy measures between experts and learners is crucial in their effectiveness in learning from demonstrations. In this paper, we present tractable solutions by formulating imitation learning as minimization of the Sinkhorn distance between occupancy measures. The formulation combines the valuable properties of optimal transport metrics in comparing non-overlapping distributions with a cosine distance cost defined in an adversarially learned feature space. This leads to a highly discriminative critic network and optimal transport plan that subsequently guide imitation learning. We evaluate the proposed approach using both the reward metric and the Sinkhorn distance metric on a number of MuJoCo experiments. For the implementation and reproducing results please refer to the following repository \url{https://github.com/gpapagiannis/sinkhorn-imitation}.
\end{abstract}

\section{Introduction}\label{sec:intro}
Recent developments in reinforcement learning (RL) have allowed agents to achieve state-of-the-art performance on complex tasks from learning to play games ~\cite{alphago,alphastar,dqn} to dexterous manipulation \cite{Rajeswaran-RSS-18}, provided with well defined reward functions. However, crafting such a reward function in practical scenarios to encapsulate the desired objective is often non-trivial. Imitation learning (IL) \cite{il-alg-perspective} aims to address this issue by formulating the problem of learning behavior through expert demonstration and has shown promises on various application domains including autonomous driving and surgical task automation \cite{helicopter-il,surgical-task,zucker2011,alvinn,koberandpeters2009}.

The main approaches to imitation learning include that of behavioral cloning (BC) and inverse reinforcement learning (IRL). BC mimics the expert’s behavior by converting the task into a supervised regression problem \cite{alvinn,sammut92}. While simple to implement, it is known to suffer from low sample efficiency and poor generalization performance due to covariate shift and high sample correlations in the expert’s trajectory \cite{il-reductions,dagger}. Algorithms such as Dataset Aggregation (DAgger) \cite{dagger} and Disturbances for Augmenting Robot Trajectories (DART) \cite{dart} alleviate this issue. However, they require constantly querying an expert for the correct actions. 

Inverse reinforcement learning instead aims to recover a reward function which is subsequently used to train the learner’s policy \cite{maxentirl,algosforirl}. IRL approaches have shown significantly better results \cite{helicopter-il,learning-for-control-il,rl-il-visual,bayesian-irl,abeel2008parkinglotnavigation,kuderer2013} including being sample efficient in terms of expert demonstration. However, IRL itself is an ill-posed problem - multiple reward functions can characterize a specific expert behavior, therefore additional constraints need to be imposed to recover a unique solution \cite{algosforirl,maxentirl,causal-maxentirl}. In addition, the alternating optimization procedure between reward recovery and policy training leads to increased computational cost. 

Adversarial imitation learning, on the other hand, bypasses the step of explicit reward inference as in IRL and directly learns a policy that matches that of an expert. Generative adversarial imitation learning (GAIL) \cite{GAIL} minimizes the Jensen-Shannon (JS) divergence between the learner’s and expert’s occupancy measures through a generative adversarial networks (GANs)-based training process. GAIL was developed as a variant of the reward regularized maximum entropy IRL framework \cite{maxentirl}, where different reward regularizers lead to different IL methods. GAIL has been extended  by various other methods aiming to improve its sample efficiency in regard to environment interaction through off-policy RL \cite{sasaki2018,gail-off-policy-blonde,gail-observation,il-off-policy-dist-matching,dac-gail}.  Recent development \cite{fairl} provides a unified probabilistic perspective to interpret different imitation learning methods as \textit{f}-divergence minimization problems and showed that the state-marginal matching objective of IRL approaches is what contributes the most to their superior performance compared to BC. While these methods have shown empirical success, they inherit the same issues from \textit{f}-divergence and adversarial training, such as training instability in GAN-based training \cite{gans} and mode-covering behavior in the JS and Kullback-Leibler (KL) divergences \cite{fairl,ke2019}.

An alternative approach is to utilize optimal transport-based metrics to formulate the imitation learning problem. The optimal transport (OT) theory \cite{villani2008} provides a flexible and powerful tool to compare probability distributions through coupling of distributions based on the metric in the underlying spaces. The Wasserstein adversarial imitation learning (WAIL) \cite{WAIL} was proposed to minimize the dual form of the Wasserstein distance between the learner’s and expert’s occupancy measures, similar to the training of the Wasserstein GAN \cite{w-gan}. The geometric property of the Wasserstein distance leads to numerical stability in training and robustness to disjoint measures. However, the solution to the dual formulation is intractable; approximations are needed in the implementation of neural networks to impose the required Lipschitz condition \cite{improved-gans}. \cite{dadashi2020} introduced Primal Wasserstein imitation learning (PWIL), that uses a reward proxy derived based on an upper bound to the Wasserstein distance between the state-action distributions of the learner and the expert. While PWIL leads to successful imitation, it is unclear how it inherits the theoretical properties of OT, since the transport map between occupancy measures is suboptimal, based on a greedy coupling strategy whose approximation error is difficult to quantify. 

In this paper we present Sinkhorn imitation learning (SIL), a \emph{tractable} solution to optimal transport-based imitation learning by leveraging the coupling of occupancy measures and the computational efficiency of the Sinkhorn distance \cite{sinkhorn}, that inherits the theoretical properties of OT. Our main contributions include: (i) We propose and justify an imitation learning training pipeline that minimizes the Sinkhorn distance between occupancy measures of the expert and the learner; (ii) We derive a reward proxy using a set of trainable and highly discriminative optimal transport ground metrics; (iii) We demonstrate through experiments on the MuJoCo simulator \cite{mujoco} that SIL obtains comparable results with the state-of-the-art, outperforming the baselines on a number of experiment settings in regard to both the commonly used reward metric and the Sinkhorn distance.

The rest of this paper is organized as follows. In Section \ref{sec:background} we provide the necessary background for this work. Section \ref{sec:sil} introduces the proposed Sinkhorn Imitation Learning (SIL) framework. Section \ref{sec:experiments} provides details of experiments to evaluate the performance of SIL on a number of MuJoCo environments. We conclude the paper and discuss future research directions in Section \ref{sec:conclusion}.

\section{Background}\label{sec:background}
\subsection{Imitation Learning}
\textbf{Notation.} We consider a Markov Decision Process (MDP) which is defined as a tuple \(\{\mathcal{S}, \mathcal{A}, \mathcal{P}, r, \gamma\}\), where \(\mathcal{S}\) is a set of states, \(\mathcal{A}\) is a set of possible actions an agent can take on the environment, \(\mathcal{P}:\mathcal{S}\times\mathcal{A}\times\mathcal{S}\rightarrow[0,1]\) is a transition probability matrix, \(r:\mathcal{S}\times\mathcal{A}\rightarrow\mathbb{R}\) is a reward function and \(\gamma\in(0, 1)\) is a discount factor. The agent's behavior is defined by a stochastic policy \(\pi:\mathcal{S}\rightarrow \textrm{Prob}(\mathcal{A})\) and \(\Pi\) is the set of all such policies. We use \(\pi_E, \pi\in\Pi\) to refer to the expert and learner policy respectively. The performance measure of policy \(\pi\) is defined as \(\mathcal{J} =\mathbb{E}_\pi[r(s, a)]=\mathbb{E}[\sum_{t=0}^\infty\gamma^tr(s_t, a_t)| \mathcal{P}, \pi]\) where \(s_t\in\mathcal{S}\) is a state observed by the agent at time step \(t\). With a slight abuse of
notations, we also use $r((s,a)_\pi)$ to denote explicitly that $(s,a)_\pi\sim\pi$.  \(\tau_E\) and \(\tau_\pi\) denote the set of state-action pairs sampled by an expert and a learner policy respectively during interaction with the environment, also referred to as trajectories. The distribution of state-action pairs generated by policy $\pi$ through environment interaction, also known as the occupancy measure \(\rho_\pi : \mathcal{S}\times\mathcal{A}\rightarrow\mathbb{R}\), is defined as \label{eq:occupancy-measure}
    \(\rho_\pi(s,a) =(1-\gamma) \pi(a|s)\overset{\infty}{\underset{t=0}{\sum} }\gamma^tP_\pi[s_t=s]\) where \(P_\pi[s_t=s]\) denotes the probability of a state being $s$ at time step $t$ following policy $\pi$.

\paragraph{Generative Adversarial Imitation Learning.} Ho and Ermon \cite{GAIL} extended the framework of MaxEnt IRL by introducing a reward regularizer \(\psi(r): \mathcal{S}\times\mathcal{A} \rightarrow \mathbb{R}\):
\begin{equation}\label{eq:irl}
    \textrm{IRL}_\psi(\pi_E):=\underset{r}{\arg\max}-\psi(r)+\underset{\pi\in\Pi}{\min}\big(-\mathcal{H}^{causal}(\pi) - \mathbb{E}_{\pi}[r(s,a)]\big) +\mathbb{E}_{\pi_E}[r(s,a)]\,\,,
\end{equation}
where \(\mathcal{H}^{causal}(\pi):=\mathbb{E}_{\rho_\pi}[-\log\pi(a|s)]/(1-\gamma)\) \cite{causal-maxentirl}. The process of RL following IRL can be formulated as that of occupancy measure matching~\cite{GAIL}:
\begin{equation}\label{eq:rl-irl}
\textrm{RL}\circ \textrm{IRL}_{\psi}(\pi_E):=\underset{\pi\in\Pi}{\arg\min}-\mathcal{H}^{causal}(\pi)+\psi^*(\rho_{\pi}-\rho_{E})\,\,,
\end{equation}
where \(\psi^*\) corresponds to the convex conjugate of the reward regularizer \(\psi(r)\). The regularized MaxEnt IRL framework bypasses the expensive step of reward inference and learns how to imitate an expert by matching its occupancy measure. Different realizations of the reward regularizer lead to different IL frameworks. A specific choice of the regularizer leads to the Generative Adversarial Imitation Learning (GAIL) framework that minimizes the Jensen-Shannon divergence between the learner's and expert's occupancy measures~\cite{GAIL}.

\paragraph{\textit{f}-Divergence MaxEnt IRL.} Recently, Ghasemipour et al. \cite{fairl} showed that training a learner policy \(\pi\) to minimize the distance between two occupancy measures can be generalised to minimize any \textit{f}-divergence between \(\rho_{E}\) and \(\rho_\pi\) denoted as \(D_f(\rho_{E}\,\, \|\,\, \rho_\pi)\). Different choices of \textit{f} yield different divergence minimization IL algorithms \cite{fairl} and can be computed as:
\begin{equation}
\max_{T_\omega}\mathbb{E}_{(s,a)\sim\rho_{E}}[T_\omega(s,a)]-\mathbb{E}_{(s,a)\sim\rho_{\pi}}[f^*(T_\omega(s,a)))]\,\,,
\end{equation}where \(T_\omega:\mathcal{S}\times\mathcal{A}\rightarrow\mathbb{R}\) and \(f^*\) is the convex conjugate of the selected \textit{f}-divergence. The learner's policy is optimized with respect to the reward proxy \(
    f^*(T_\omega(s, a))\).

\subsection{Optimal Transport}
While divergence minimization methods have enjoyed empirical success, they are still difficult to evaluate in high dimensions \cite{integral-probability-metrics}, due to the sensitivity to different hyperparameters and difficulty in training depending on the distributions that are evaluated \cite{improved-gans}. The optimal transport (OT) theory \cite{villani2008} provides effective methods to compare degenerate distributions by accounting for the underlying metric space. Consider \(P_k(\Gamma)\) to be the set of Borel probability measures on a Polish metric space \((\Gamma,d)\) with finite $k$-th moment. Given two probability measures $p,q\in P_k(\Gamma)$, the $k$-Wasserstein metric is defined as~\cite{villani2008}:
\begin{equation}\label{eq:primal-wasserstein}
    \mathcal{W}_k(p, q)_c=\Big(\inf_{\zeta\in\Omega(p,q)}\int_\Gamma c(x, y)^kd\zeta(x, y)\Big)^\frac{1}{k}\,\,,
\end{equation}
where \(\Omega(p,q)\) denotes the set of joint probability distributions whose marginals are $p$ and $q$, respectively. \(c(x,y)\) denotes the cost of transporting sample \(x\sim p\) to \(y\sim q\). The joint distribution \(\zeta\) that minimizes the total transportation cost is referred to as the optimal transport plan.

\paragraph{Sinkhorn Distances.} The solution to Equation (\ref{eq:primal-wasserstein}) is generally intractable for high dimensional distributions in practice. A regularized form of the optimal transport formulation was proposed by Cuturi \cite{sinkhorn} that can efficiently compute the Wasserstein metric. The Sinkhorn distance $\mathcal{W}_s^\beta(p,q)_c$ between $p$ and $q$ is defined as:
\begin{gather}
    \label{eq:sinkhorn-distannce}
    \mathcal{W}^\beta_{\textrm{s}}(p, q)_c=\underset{\zeta_\beta\in\Omega_{\beta}(p, q)}\inf\mathbb{E}_{x,y\sim\zeta_\beta}[c(x,y)]\,\,,
\end{gather}
where $\Omega_\beta(p,q)$ denotes the set of all joint distributions in $\Omega(p,q)$ with entropy of at least $\mathcal{H}(p)+\mathcal{H}(q)-\beta$ and $\mathcal{H}(\cdot)$ computes the entropy of a distribution.  The distance is evaluated on two distributions $p$ and $q$ where in the context of adversarial IL correspond to the state-action distributions of the learner and the expert policies.

\section{SIL: Sinkhorn Imitation Learning}\label{sec:sil}

We consider the problem of training a learner policy \(\pi\) to imitate an expert, by matching its state-action distribution \(\rho_{E}\) in terms of minimizing their Sinkhorn distance. To facilitate the development of the learning pipeline, we begin by discussing how the Sinkhorn distance is used to evaluate similarity between occupancy measures.

Consider the case of a learner \(\pi\) interacting with an environment and generating a trajectory of state-action pairs \(\tau_\pi\sim\pi\) that characterizes its occupancy measure. A trajectory of expert demonstrations \(\tau_{E}\sim\pi_E\) is also available as the expert trajectories. The optimal transport plan \(\zeta_\beta\) between the samples of \(\tau_\pi\) and \(\tau_E\) can be obtained via the Sinkhorn algorithm \cite{sinkhorn}. Following Equation (\ref{eq:sinkhorn-distannce}) we can evaluate the Sinkhorn distance of \(\tau_\pi\) and \(\tau_E\) as follows:
\begin{equation}\label{eq:sinkhorn-empirical}
    \mathcal{W}^\beta_{\textrm{s}}(\tau_\pi, \tau_{E})_c=\sum_{(s,a)_\pi\in \tau_\pi}\sum_{(s,a)_{\pi_E}\in\tau_E}c\Big((s,a)_\pi,(s,a)_{\pi_E}\Big)\zeta_\beta\Big((s,a)_\pi, (s,a)_{\pi_E}\Big)\,\,.
\end{equation}

\noindent\textbf{Reward Proxy.} We now introduce a reward proxy suitable for training a learner policy that minimizes
$\mathcal{W}^\beta_{\textrm{s}}(\tau_\pi, \tau_{E})_c$ in order to match the expert's occupancy measure.

The reward function $v_c((s,a)_{\pi})$ for each sample $(s,a)_\pi$ in the learner's trajectory is defined as:
\begin{equation}\label{eq:sinkhorn-reward}
    v_c((s,a)_{\pi}):=-\sum_{(s,a)_{\pi_E}\in\tau_E}c\Big((s,a)_\pi,(s,a)_{\pi_E}\Big)\zeta_\beta\Big((s,a)_\pi, (s,a)_{\pi_E}\Big)\,\,.
\end{equation}

\noindent The optimization objective of the learner policy $\mathcal{J}=\mathbb{E}_{\pi}[r((s,a)_\pi)]$ under $r((s,a)_\pi):=v_c((s,a)_\pi)$ corresponds to minimizing the Sinkhorn distance between the learner's and expert's trajectories defined in Equation (\ref{eq:sinkhorn-empirical}). Hence, by maximizing the optimization objective $\mathcal{J}$ with reward $v_c((s,a)_\pi)$, a learner is trained to minimize the Sinkhorn distance between the occupancy measures of the learner and the expert demonstrator.\\

\noindent\textbf{Adversarial reward proxy.} The reward specified in Equation (\ref{eq:sinkhorn-reward}) can only be obtained after the learner has generated a complete trajectory. The optimal transport plan \(\zeta_\beta\big((s,a)_{\pi}, (s,a)_{\pi_E}\big)\) then weighs the transport cost of each sample \((s,a)_\pi\in\tau_\pi\) according to the samples present in \(\tau_\pi\) and \(\tau_E\). The dependence of \(v_c((s,a)_\pi)\) to all state-action pairs in \(\tau_\pi\) and \(\tau_E\) can potentially result in the same state-action pair being assigned significantly different rewards depending on the trajectory that it is sampled from.
Such dependence can lead to difficulty in maximizing the optimization objective $\mathcal{J}$ (and equivalently in minimizing the Sinkhorn distance between the occupancy measures from the learner and the expert). Empirical evidence is provided in the ablation study in Section \ref{sec:experiments}.

In order to provide a discriminative signal to the learner's policy and aid the optimization process, we consider adversarially training a critic to penalize non-expert state-action pairs by increasing their transport cost to the expert’s distribution, drawing inspiration from the adversarially trained transport ground metric in the OT-GAN framework \cite{ot-gan}. The critic \(c_w((s,a)_\pi, (s,a)_{\pi_E})\) parameterized by \(w\) is defined as follows:
\begin{equation}\label{eq:transport-cost}
        c_w((s,a)_\pi, (s,a)_{\pi_E})=1-\frac{f_w((s,a)_\pi) \cdot f_w((s,a)_{\pi_E})}{||f_w((s,a)_\pi)||_2||f_w((s,a)_{\pi_E})||_2}\,\,,
\end{equation}
where $\cdot$ denotes the inner product between two vectors. $f_w(\cdot): \mathcal{S}\times\mathcal{A}\rightarrow\mathbb{R}^d$ maps the environment's observation space $\mathcal{S}\times\mathcal{A}$ to an adversarially learned feature space $\mathbb{R}^d$ where $d$ is the feature dimension.
The adversarial reward proxy $v_{c_w}((s,a)_\pi)$ is obtained by substituting the transport cost $c(\cdot,\cdot)$ in Equation~\eqref{eq:sinkhorn-reward} with $c_w(\cdot,\cdot)$ defined by Equation~\eqref{eq:transport-cost}. SIL learns $\pi$
by solving the following minimax optimization problem:
\begin{equation}
   \underset{\pi}{\arg\min}\,\, \max_w\mathcal{W}_s^\beta(\rho_\pi,\rho_E)_{c_w}\,\,.
\end{equation}\vspace{
-5mm}
\begin{remark}For SIL, the adversarial training part of the transport cost is not part of the approximation procedure of the distance metric, as in GAIL~\cite{GAIL} and WAIL~\cite{WAIL}. The Sinkhorn distance is computed directly via the Sinkhorn iterative procedure \cite{sinkhorn} with the transport cost defined in Equation \eqref{eq:transport-cost}.
\end{remark}

\noindent\textbf{Algorithm.} The pseudocode for the proposed Sinkhorn imitation learning (SIL) framework is presented in Algorithm \ref{algo:SIL}. In each iteration we randomly match each of the learner's generated trajectories to one of the expert's and obtain their Sinkhorn distance. The reason behind this implementation choice is to maintain a constant computational complexity with respect to a potentially increasing number of demonstrations. We then alternate between one step of updating a critic network \(c_w\) to maximize the Sinkhorn distance between the learner's and expert's trajectories and a policy update step to minimize the distance between occupancy measures with the learned reward proxy. As SIL depends on complete environment trajectories to compute the Sinkhorn distance, it is inherently an on-policy method. Hence, to train our imitator policy we use Trust Region Policy Optimization (TRPO) \cite{trpo} for our experiments.



\begin{algorithm}[t!]
\SetAlgoLined
\KwIn{Set of expert trajectories \(\{\tau_E\}\sim \pi_E\), Sinkhorn regularization parameter \(\beta\), initial learner's policy parameters \(\theta_0\), initial critic network parameters \(w_0\), number of training iterations \(K\)}
\begin{algorithmic}[1]
\FOR{\(\mbox{iteration } k=0 \textrm{ to } K\)}
\STATE Sample a set of trajectories \(\{\tau_{\pi_{\theta_k}}\}_k\sim\pi_{\theta_k}\).
\STATE Create a set of trajectory pairs \(\{(\tau_{\pi_{\theta_k}}, \tau_E)\}_k\) by randomly \\matching trajectories from the learner's set to the expert's.
\STATE For each pair in \(\{(\tau_{\pi_{\theta_k}}, \tau_E)\}_k\), calculate \(\mathcal{W}^{\beta}_{s}(\tau_{\pi_{\theta_k}}, \tau_{E})_{c_w}\) using the Sinkhorn \\algorithm (Equation~\eqref{eq:sinkhorn-distannce}) and transport cost as in Equation (\ref{eq:transport-cost}), in order to update the \\reward proxy \(v_{c_{w_k}}((s,a)_{\pi_{\theta_k}})\)
for each state action pair.
\STATE Update \(w_k\) to maximize $\mathcal{W}^{\beta}_{s}(\tau_{\pi_{\theta_k}}, \tau_{E})_{c_w}$ using gradient ascent with the gradient:
\begin{equation}\label{eq:sinkhorn-update}
   \nabla_{w_k}\frac{1}{m}\sum_{ \{(\tau_{\pi_{\theta_k}}, \tau_E)\}_k }\mathcal{W}^{\beta}_{s}(\tau_{\pi_{\theta_k}}, \tau_{E})_{c_w}\,\,,
\end{equation}where \(m\) is the number of trajectory pairs.
\STATE Update policy parameter \(\theta_k\) using TRPO and reward  \(v_{c_{w_k}}((s,a)_{\pi_{\theta_k}})\) updated in Step 4.
\ENDFOR
\caption{Sinkhorn imitation learning (SIL)}\label{algo:SIL}
\end{algorithmic}{}
\KwOut{Learned policy \(\pi_{\theta_k}\).}
\end{algorithm}


\subsection{Connection to regularized MaxEnt IRL.}\label{sec:connection}
We now show how SIL can be interpreted as a variant of the regularized MaxEnt IRL framework \cite{GAIL} 
given a specific choice of \(\psi(r)\).

\begin{definition}
Consider a learner's policy and expert's demonstrations, as well as their induced occupancy measures \(\rho_\pi\) and \(\rho_E\). We define the following reward regularizer:
\begin{equation}
    \psi_{\mathcal{W}}(r) := - \mathcal{W}_{s}^{\beta}(\rho_\pi, \rho_{E})_{c_w} + \mathbb{E}_{\rho_\pi}[r(s,a)] - \mathbb{E}_{\rho_E}[r(s,a)]\,\,.
    \label{eq:regularizer}
\end{equation}
\end{definition}

\begin{proposition}
The reward regularizer \(\psi_\mathcal{W}(r)\) defined in Equation~\eqref{eq:regularizer} leads to an entropy regularized MaxEnt IRL algorithm. When \(r((s,a)_\pi)=v_{c_w}((s,a)_\pi)\),

\begin{equation}
    \textrm{RL}\circ \textrm{IRL}_{\psi_{\mathcal{W}}}(\pi_E)=\underset{\pi\in\Pi}{\arg\min}-\mathcal{H}^{causal}(\pi)+\sup_{w}\mathcal{W}_{s}^\beta(\rho_\pi, \rho_{E})_{c_w}\,\,.
\label{eq:sinkhorn_maxEntIRL}
\end{equation} 
\label{prop:sinkhorn_maxEntIRL}
\end{proposition}


Equation~\eqref{eq:sinkhorn_maxEntIRL} corresponds to the process of updating a critic network to maximize the Sinkhorn distance between the learner's and expert's occupancy measures, followed by the process of finding a policy $\pi$ to minimize it. The added term $\mathcal{H}^{causal}(\pi)$ is treated as a regularization parameter. 
\begin{proof} Consider the set of possible rewards \(\mathcal{R}:=\{r:\mathcal{S}\times\mathcal{A}\rightarrow\mathbb{R}\}\) in finite state-action space
as in~\cite{GAIL} and~\cite{fairl}. The joint state-action distributions $\rho_\pi$ and $\rho_E$ are represented as vectors in $[0, 1]^{\mathcal{S}\times\mathcal{A}}$.

Define $\psi_{\mathcal{W}}(r) := - \mathcal{W}_{s}^{\beta}(\rho_\pi, \rho_{E}) + \mathbb{E}_{\rho_\pi}[r(s,a)] - \mathbb{E}_{\rho_E}[r(s,a)]$, where $\mathcal{W}_{s}^{\beta}(\rho_\pi, \rho_{E})$ is obtained with the transport cost $c_w$ defined in Equation \eqref{eq:transport-cost}. Given $r(s,a)=v_{c_w}(s,a)$ and recall that the convex conjugate of a function \(g\) is \(g^*(y)=\sup_{x\in\textrm{dom}(g)}(y^Tx-g(x))\),
we obtain

\begin{align}
\psi_{\mathcal{W}}^*(\rho_\pi-\rho_{E}) =&\sup_{r\in\mathcal{R}}[(\rho_\pi-\rho_{E})^T r-\psi_{\mathcal{W}}(r)]=\sup_{r\in\mathcal{R}}[\sum_{\mathcal{S\times A}}(\rho_\pi(s,a)-\rho_{E}(s,a)) \cdot r(s,a) \nonumber\\
&+ \mathcal{W}_{s}^{\beta}(\rho_\pi, \rho_{E})- \sum_{\mathcal{S\times A}}(\rho_\pi(s,a)-\rho_{E}(s,a)) \cdot r(s,a)] =\nonumber\\
&\sup_{r\in\mathcal{R}}\,\,\mathcal{W}_{s}^{\beta}(\rho_\pi,\rho_{E})
=\sup_{v_{c_w}\in\mathcal{R}}\,\,\mathcal{W}_{s}^{\beta}(\rho_\pi, \rho_{E})=\sup_{w} \mathcal{W}_{s}^{\beta}(\rho_\pi, \rho_{E})\,\,.
\end{align}

From Equation~\eqref{eq:rl-irl},
\begin{align}
\textrm{RL}\circ \textrm{IRL}_{\psi}(\pi_E) &=\underset{\pi\in\Pi}{\arg\min}-\mathcal{H}^{causal}(\pi)+\psi_{\mathcal{W}}^*(\rho_{\pi}-\rho_{E})\nonumber\\
&=\underset{\pi\in\Pi}{\arg\min}-\mathcal{H}^{causal}(\pi)+\sup_{w}\,\, \mathcal{W}_{s}^{\beta}(\rho_\pi, \rho_{E})\,\,.
\end{align}

\end{proof}
\section{Experiments}\label{sec:experiments}
To empirically evaluate the Sinkhorn imitation learning (SIL) algorithm, we benchmark SIL against BC in the four MuJoCo \cite{mujoco} environments studied in \cite{fairl}, namely Hopper-v2, Walker2d-v2, Ant-v2 and HalfCheetah-v2, as well as the Humanoid-v2 environment. Given that SIL is an on-policy method due to the requirement of complete trajectories, two on-policy adversarial IL algorithms, namely GAIL \cite{GAIL} and AIRL \cite{airl}, are also included as baselines. All algorithms are evaluated against the true reward metric obtained through environment interaction, in addition to the Sinkhorn distance between
the samples from the learned policy and the expert demonstrations.

Initially we train policies using TRPO \cite{trpo} to obtain expert performance. The expert policies are used to generate sets of expert demonstrations. The performance of the obtained expert policies can be found in Table \ref{tab:exp-per}. To study the robustness of SIL in learning from various lengths of trajectory sets we train the algorithms on sets of \(\{2,4,8,16,32\}\) and for Humanoid-v2 for \(\{8,16,32\}\) sets. All trajectories are subsampled by a factor of $20$ starting from a random offset, a common practice found in \cite{GAIL,fairl,airl}. SIL, GAIL and AIRL are trained for 250 iterations allowing approximately $50,000$ environment interactions per iteration. For Humanoid-v2 we train the algorithms for $350$ iterations. All reported results correspond to performance metrics obtained after testing the learner policies on $50$ episodes.
\begin{table}[h!]
  \begin{center}
    \small
    \resizebox{0.4\textwidth}{!}{
    \begin{tabular}{lc}
    \Xhline{1\arrayrulewidth}
      \textbf{Environments}  &\textbf{Expert Performance} \TBstrut\TBstrut \\
    \Xhline{0.5\arrayrulewidth}
      Hopper-v2             &\(3354.74 \pm1.87\)\Tstrut \Bstrut\\
                   
         HalfCheetah-v2         &  \(4726.53 \pm 133.12 \)\Tstrut \Bstrut\\
                  
         Walker2d-v2     &     \(3496.44 \pm 8.79\) \Tstrut \Bstrut\\
                    
    Ant-v2       &   \(5063.11 \pm 337.50\)\Tstrut \Bstrut\\

      Humanoid-v2           &  \(6303.36 \pm 97.71\) \Tstrut \Bstrut\\
                    
    \Xhline{1\arrayrulewidth}
 \end{tabular}}
  \end{center}
      \caption{Performance of expert policies providing the demonstrations trained using TRPO. \label{tab:exp-per}}
\end{table}
\subsection{Implementation Details}\label{sec:impl-det}

\textbf{Adversarial Critic. }The critic network consists of a 2-layer MLP architecture with 128 units each with ReLU activations.  For each experiment we report the best performing result after training the critic with the following learning rates $\{0.0004, 0.0005, 0.0006, \allowbreak 0.0007,  0.0008, 0.0009\}$ and output dimensions $\{5,10,30\}$. 
Although different choices of the critic network output dimension may yield better results for the proposed SIL algorithm in different environments, no further attempt was made to fine-tune the output for the critic. We note that for most experiment settings a critic output dimension of $30$ and learning rate of $0.0005$ among the pool of candidate values yield the best results.\\\\
\textbf{Reward Proxy. }After obtaining the value of $v_{c_w}$ as defined in Equations  \eqref{eq:sinkhorn-reward} and  \eqref{eq:transport-cost}, we add a value of $\frac{2}{L}$ where $L$ is the trajectory length and scale the reward by $2$. By doing so we set the range of $v_{c_w}$ to be $0\leq v_{c_w}\leq4$ which proved to be effective for environments requiring a survival bonus. We keep track of a running standard deviation to normalize rewards.
\\\\\textbf{Policy Architecture \& Training. }For both the expert and learner policies, we use the same architecture comprised of a 2-layer MLP architecture each with 128 units with ReLU activations. The same architecture is used amongst all imitation learning algorithms. For all adversarial IL algorithms, as well as obtaining expert performance, we train the policies using Trust Region Policy Optimization \cite{trpo}.
Finally, we normalize environment observations by keeping track of the running mean and standard deviation.
\\\\\textbf{GAIL \& AIRL. }
To aid the performance of the benchmarks algorithms GAIL and AIRL in the HalfCheetah-v2 environment, we initialize the policies with that from behavioural cloning.
\\\\\textbf{Computational Resource. } The experiments were run on a computer with an Intel (R) Xeon (R) Gold 5218 CPU 2.3 GHz and 16GB of RAM, and a RTX 6000 graphic card with 22GB memories.
\subsection{Results}
\textbf{Sinkhorn metric.} We begin by evaluating performance amongst IL methods using the Sinkhorn metric. Since our goal is to assess how well imitation learning algorithms match the expert's occupancy measure, the Sinkhorn distance offers a valid metric of similarity between learner’s and expert’s trajectories compared to the reward metric which is also often unavailable in practical scenarios.
We report the Sinkhorn distance between occupancy measures computed with a fixed cosine distance-based transport cost during testing and evaluation:
\begin{equation}\label{eq:fixed-cosine}
      c((s,a)_\pi, (s,a)_{\pi_E})=1-\frac{[s,a]_\pi \cdot [s,a]_{\pi_E}}{||[s,a]_\pi||_2||[s,a]_{\pi_E}||_2}\,\,,
\end{equation}
where $[s,a]_\pi$ denotes the concatenated vector of state-action of policy $\pi$ and $||\cdot||_2$ computes the L2 norm.
\begin{figure}[t!]
    \centering
    \includegraphics[width=\textwidth]{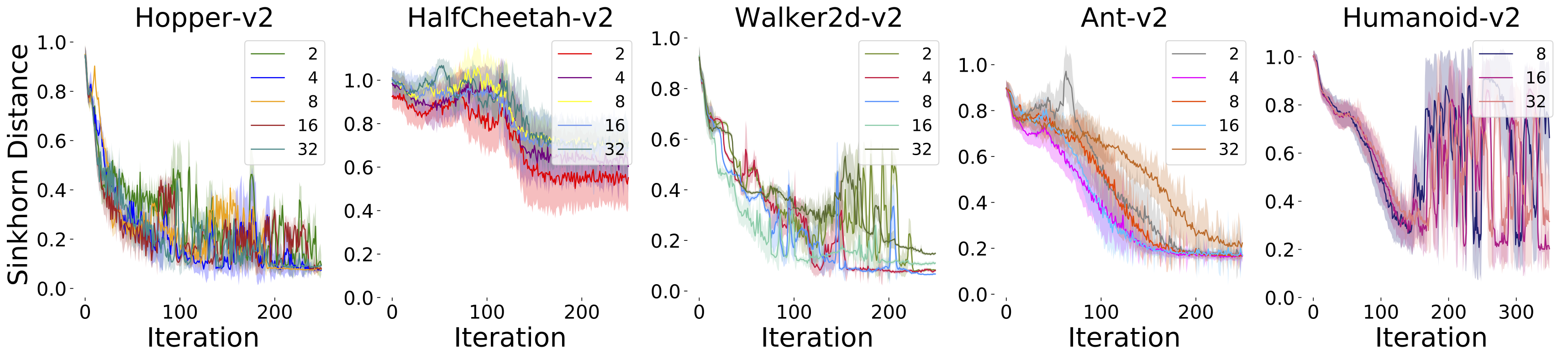}
    \caption{Mean and standard deviation of the Sinkhorn distance evaluated during training of SIL using a fixed cosine transport cost by stochastically sampling an action from the learner's policy.}
    \label{fig:sil-subsampled-training}
\end{figure}{}
\begin{table}[t!]
  \begin{center}
   \centering

\resizebox{\columnwidth}{!}{ \begin{tabular}{lccccc} 
    \Xhline{2\arrayrulewidth}
      \textbf{Environments} & \textbf{Trajectories} & \textbf{BC} & \textbf{GAIL} & \textbf{AIRL} & \textbf{SIL} \TBstrut \\
    \Xhline{1\arrayrulewidth}
                   &2     & \(0.467\pm0.009\)       & \(0.098\pm0.003\) & \(\mathbf{0.069\pm0.001}\) & \(0.073\pm0.001\)\Tstrut \Bstrut\\
    Hopper-v2      &4     & \(0.408\pm0.080\)       & \(0.120\pm0.010\) & \(\mathbf{0.066\pm0.009}\)& \(0.082\pm0.010\) \Tstrut \Bstrut\\
                   &8     & \(0.300\pm0.029\)      & \( 0.074\pm0.004\) & \(\mathbf{0.068\pm0.006}\)  & \(0.071\pm0.005\)\Tstrut \Bstrut\\
         
                   &16    & \(0.182\pm0.042\)        & \( 0.106\pm0.008\) &\(\mathbf{0.074\pm0.010}\) &\(0.078\pm 0.012\)\Tstrut \Bstrut\\
                   &32    & \(0.157\pm0.084\)        & \(\mathbf{0.071\pm0.008}\) &\(0.072\pm0.009\) &\(0.089\pm0.008\)\Tstrut \Bstrut\\

    \Xhline{0.5\arrayrulewidth}

                        &2   & \(1.043\pm0.058\)    & \(0.940\pm0.181\)& \(0.577\pm0.157\) & \( \mathbf{0.546\pm0.138}\)\Tstrut \Bstrut\\
    HalfCheetah-v2      &4   & \(0.791\pm0.096\)    & \(0.633\pm0.095\)& \(0.630\pm0.091\) & \(\mathbf{ 0.620\pm0.101}\)\Tstrut \Bstrut\\
                        &8   & \(0.841\pm0.071\)     & \(0.702\pm0.095\)& \(0.708\pm0.054\)& \(\mathbf{0.700\pm0.052}\) \Tstrut \Bstrut\\
                        &16  & \(0.764\pm0.166\)      & \(\mathbf{0.670\pm0.128}\)       &\(0.671\pm0.112\)&\(0.688\pm0.131\) \Tstrut \Bstrut\\
                        &32    & \(0.717\pm0.129\)  & \(0.695\pm0.113\) &\(0.699\pm0.091\)&\(\mathbf{0.685\pm0.083}\) \Tstrut \Bstrut\\

    \Xhline{0.5\arrayrulewidth}
                        &2    & \(0.474\pm0.023\)       & \(0.067\pm0.008\)& \(\mathbf{0.034\pm0.005}\)  &\(0.080\pm0.004\) \Tstrut \Bstrut\\
    Walker2d-v2         &4   & \(0.694\pm0.011\)        & \(0.067\pm0.006\)& \(\mathbf{0.036\pm0002}\)  &\( 0.079\pm0.005\)\Tstrut \Bstrut\\
                        &8       & \(0.335\pm0.004\)         & \(0.069\pm0.005\)& \(\mathbf{0.036\pm0.003}\) & \(0.063\pm0.003\) \Tstrut \Bstrut\\
                        &16    & \(0.199\pm0.013\)        & \(0.061\pm0.004\) &\(\mathbf{0.037\pm0.005}\) &\(0.102\pm0.007\)\Tstrut \Bstrut\\
                        &32    & \(0.196\pm0.098\)        & \(0.052\pm0.003\) &\(\mathbf{0.042\pm0.004}\) &\(0.147\pm0.003\)\Tstrut \Bstrut\\

        \Xhline{0.5\arrayrulewidth}
                        &2     & \(0.843\pm0.033\)        & \(0.344\pm0.068\) & \(0.164\pm0.006\)&\(\mathbf{0.158\pm0.008}\) \Tstrut \Bstrut\\
    Ant-v2              &4     & \(0.684\pm0.159\)        & \(0.165\pm0.119\) & \(0.163\pm0.008\) &\(\mathbf{0.157\pm0.014}\)\Tstrut \Bstrut\\
                        &8     & \(0.996\pm0.029\)        & \(0.159\pm0.016\) & \(0.164\pm0.019\)&\(\mathbf{0.155\pm0.012}\)\Tstrut \Bstrut\\
                        &16    & \( 0.724\pm0.149\)       & \(0.225\pm0.106\) &\(0.173\pm0.062\) &\(\mathbf{0.165\pm0.022}\)\Tstrut \Bstrut\\
                        &32    & \(0.452\pm0094\)         & \(0.176\pm0.029\) &\(\mathbf{0.172\pm0.020}\)   &\(0.173\pm0.018\)                       \Tstrut \Bstrut\\

        \Xhline{1\arrayrulewidth}
 \TBstrut\Tstrut   &8&\(\mathbf{0.336\pm0.089}\)&\(0.386\pm0.011\)&\(1.015\pm0.015\)&\(0.379\pm0.296\)  \TBstrut\Tstrut\\
  \TBstrut\Tstrut  Humanoid-v2&16&\(0.290\pm0.086\)&\(0.428\pm0.027\)&\(1.034\pm0.017\)&\(\mathbf{0.182\pm0.011}\)  \TBstrut\Tstrut\\
  \TBstrut\Tstrut  &32&\(0.182\pm0.028\)&\(\mathbf{0.162\pm0.144}\)&\(1.026\pm0.015\)&\(0.250\pm0.180\)  \TBstrut\Tstrut\\
        \Xhline{2\arrayrulewidth}

    \end{tabular}}         
   
  \end{center}

   \caption{Mean and standard deviation of the Sinkhorn distance between the expert demonstrations and samples from imitator policies for BC, GAIL, AIRL and SIL. A fixed cosine transport cost is used only for evaluation (Smaller distance denotes better performance). \label{tab:il-sinkhorn-policies}}
   \vspace{-10mm}
\end{table}
Table \ref{tab:il-sinkhorn-policies} reports the Sinkhorn metric evaluated between the trajectories generated by the learned policies with the demonstrations provided by the expert. A smaller Sinkhorn distance corresponds to higher similarity between the learner's and expert's generated trajectories. SIL, AIRL and GAIL obtain comparable performance in most of the environments. The proposed SIL algorithm outperforms the baselines in almost all experiments on the environments of HalfCheetah-v2 and Ant-v2, while AIRL achieves superior performance on the environments of Hopper-v2 and Walker2d-v2. GAIL on the other hand obtains relatively poor performance with regard to the Sinkhorn distance when provided with only 2 expert trajectories on the environments of Hopper-v2, HalfCheetah-v2 and Ant-v2. As expected, behavioral cloning fails to obtain competitive performance in almost all experiment settings especially when provided with a small number of expert demonstrations.

In addition, SIL outperforms GAIL and AIRL on the Humanoid-v2 environment when provided with 8 and 16 trajectories, where SIL demonstrates significantly improved sample efficiency in terms of both expert demonstrations and environment interactions. GAIL outperforms the rest when trained with 32 trajectories on the Humanoid-v2 environment. Interestingly, BC obtains superior performance with regard to the Sinkhorn distance on the Humanoid-v2 environment when provided with 8 trajectories, but low performance regarding the reward metric as shown in Table~\ref{tab:il-policies}.\\\\ 
\textbf{Reward metric.} To better understand how performance changes in terms of the Sinkhorn distance metric translates to the true reward, Table \ref{tab:il-policies} shows the reward obtained with the learned policies in the same experiments reported in Table~\ref{tab:il-sinkhorn-policies}. While all adversarial imitation learning algorithms exhibit similar reward values compared to the expert policies, we observe that SIL generally obtains lower reward compared to AIRL on Ant-v2. In addition, AIRL obtains lower reward compared to SIL and GAIL on Walker2d-v2. However, both SIL and AIRL yield superior performance in these environments when evaluated using the Sinkhorn distance as shown in Table~\ref{tab:il-sinkhorn-policies}. The result suggests that evaluating the performance of imitation learning algorithms with a true similarity metric, such as the Sinkhorn distance, can be more reliable since our objective is to match state-action distributions.

\begin{table}[t!]
  \begin{center}
    
    \centering
    \resizebox{\columnwidth}{!}{
    \begin{tabular}{lcccccc}
    \Xhline{3\arrayrulewidth}
      \textbf{Environments} & \textbf{Trajectories} &\textbf{BC} & \textbf{GAIL} & \textbf{AIRL} & \textbf{SIL} \TBstrut\TBstrut \\
\hline
                    & 2                         & \(391.38\pm42.98\)               & \(  3341.27\pm38.96\) & \(3353.33\pm2.05\) & \(\mathbf{3376.70\pm2.45}\)\Tstrut \Bstrut\\
         & 4 &      \(659.51\pm166.32\)             & \( 3206.85\pm1.56\)   & \(\mathbf{3353.75\pm1.67}\)  & \(3325.66\pm4.24\)\Tstrut \Bstrut\\
                     Hopper-v2& 8                         & \(1094.39\pm145.93\)              & \( 3216.93\pm3.08\)   & \(\mathbf{3369.17\pm3.04}\) & \(3335.31\pm2.66\)\Tstrut \Bstrut\\
                    &16                         & \(2003.71\pm655.85\)            & \(\mathbf{3380.97\pm2.16}\)    &\(3338.07\pm2.14\) &\(3376.55\pm2.65\)  \Tstrut \Bstrut\\
                    &32  &   \(2330.82\pm1013.71\)       & \(3333.93\pm1.47\) &\(\mathbf{3361.56\pm1.93}\)  &\(3326.52\pm3.62\)\Tstrut \Bstrut\\

    \Xhline{1\arrayrulewidth}
                  
                    &2                          & \(-60.80\pm23.12\)                        & \(764.91\pm546.47\)   & \(4467.83\pm61.13\) & \( \mathbf{4664.65\pm91.73}\) \Tstrut \Bstrut\\
     
     &4&      \( 1018.68\pm236.13\)                      & \(\mathbf{5183.67\pm118.74}\)  & \(4578.84\pm102.92\) & \( 4505.88\pm130.50\)\Tstrut \Bstrut\\
                                                HalfCheetah-v2 &8& \(1590.73\pm 279.05\)                   & \(\mathbf{4902.46\pm721.43}\)  & \(4686.22\pm147.89\)& \(4818.82\pm251.27\)  \Tstrut \Bstrut\\
                                                &16    & \(2434.30\pm 733.29\)               & \(4519.49\pm157.99\)            &\(\mathbf{4783.79\pm197.27}\) &\(4492.37\pm134.35\) \Tstrut \Bstrut\\
                   &32  &   \(3598.98\pm558.70\)        & \(4661.17\pm147.21\) &\(4633.48\pm116.89\)&\(\mathbf{4795.68\pm191.90}\) \Tstrut \Bstrut\\

    \Xhline{1\arrayrulewidth}
                    &2                          &   \(591.92\pm32.77\)       & \(3509.37\pm8.08\)& \(3497.80\pm9.64\)  &\(\mathbf{3566.32\pm16.11}\) \Tstrut \Bstrut\\
       &4 &    \(314.77\pm9.21\)        & \(\mathbf{3537.63\pm4.14}\)& \(3496.61\pm10.94\) &\( 3523.73\pm21.91\) \Tstrut \Bstrut\\

                    Walker2d-v2 &8                           & \(808.37\pm5.28\)        & \(3394.15\pm4.74\)& \( \mathbf{3488.68\pm10.67}\)& \(3420.13\pm16.38\)   \Tstrut \Bstrut\\
                   &16                              & \(1281.80\pm81.11\)       & \(3444.96\pm23.99\) &\(3459.84\pm8.25\) &\(\mathbf{3557.51\pm11.67}\) \Tstrut \Bstrut\\
                   &32  &   \(1804.74\pm1154.36\)      & \(3427.61\pm9.79\) &\(\mathbf{3495.04\pm17.18}\)  &\( 3203.32\pm23.65\) \Tstrut \Bstrut\\

    \Xhline{1\arrayrulewidth}
                &2                          & \(845.14\pm172.37\)       & \(3443.87\pm716.61\) & \(\mathbf{5190.89\pm67.94}\)  &\( 4981.70\pm50.89\)\Tstrut \Bstrut\\
 
        &4&    \(897.54\pm2.14\)       & \(4912.92\pm606.99\) & \(\mathbf{5182.42\pm65.70}\)  &\(5020.71\pm89.74\)\Tstrut \Bstrut\\
                Ant-v2 &8                          &\(991.92\pm2.92\)        & \(5112.21\pm102.23\) & \(5083.30\pm77.48\) &\(\mathbf{5112.55\pm62.87}\)\Tstrut \Bstrut\\
                &16                         & \(1014.14\pm447.66\)       & \(4854.87\pm895.63\) &\(\mathbf{5034.80\pm331.64}\)  &\(4935.33\pm87.15\)\Tstrut \Bstrut\\
                                   &32  &   \(2197.20\pm487.00\)        & \(5009.60\pm247.43\) &\(\mathbf{5013.36\pm119.12}\)&\(4581.27\pm123.75\) \Tstrut \Bstrut\\

    \Xhline{0.5\arrayrulewidth}
 \TBstrut\Tstrut   &8&\(1462.47\pm1139.19\)&\(1249.26\pm187.71\)&\(3897.47\pm1047.03\)&\(\mathbf{4456.09\pm2707.92}\)  \TBstrut\Tstrut\\
  \TBstrut\Tstrut  Humanoid-v2&16&\(2100.93\pm1116.79\)&\(496.11\pm113.28\)&\(4396.01\pm433.63\)&\(\mathbf{6380.37\pm40.35}\)  \TBstrut\Tstrut\\
  \TBstrut\Tstrut  &32&\(4807.86\pm1903.08\)&\(\mathbf{6252.73\pm570.72}\)&\(1884.92\pm764.89\)&\(5593.19\pm1967.86\)  \TBstrut\Tstrut\\
    \Xhline{2\arrayrulewidth}

 \end{tabular}}
  \end{center}
      \caption{Mean and standard deviation of the reward metric performance of imitator policies for BC, GAIL, AIRL and SIL. \label{tab:il-policies}}
    \vspace{-10mm}
\end{table}\noindent\textbf{Training Stability. } Table~\ref{tab:il-sinkhorn-policies} showcases that SIL consistently minimizes the Sinkhorn distance while being robust to varying lengths of expert demonstrations. Figure \ref{fig:sil-subsampled-training} depicts the evolution of the Sinkhorn distance between occupancy measures of the learner and the expert in the training process of SIL. 
In spite of the training instability observed on Walker2d-v2 with 2 or 32 expert trajectories on the Humanoid-v2 environment, SIL still successfully learns to imitate the expert demonstrator. We speculate that training stability could be improved in these settings with further hyperparameter tuning as discussed in Section \ref{sec:conclusion} which we leave for future work. Training stability of SIL is evident on the Hopper-v2, Ant-v2 and HalfCheetah-v2 environments.\\\\\noindent\textbf{Ablation study.} To study the effect of minimizing the Sinkhorn distance between occupancy measures using a fixed transport cost, we repeat our experiments on the environments Hopper-v2, HalfCheetah-v2, Walker2d-v2 and Ant-v2 with \(\{2,8,32\}\) trajectory sets. For Humanoid-v2 we conduct the experiments on sets of $\{8,16,32\}$.  In this ablation study, instead of training a critic network in an adversarially learned feature space, we assign a reward proxy defined by Equation \eqref{eq:sinkhorn-reward} with a \textit{fixed} cosine transport cost introduced in Equation \eqref{eq:fixed-cosine}.
\begin{figure}[t!]
    \centering
    \includegraphics[width=\columnwidth]{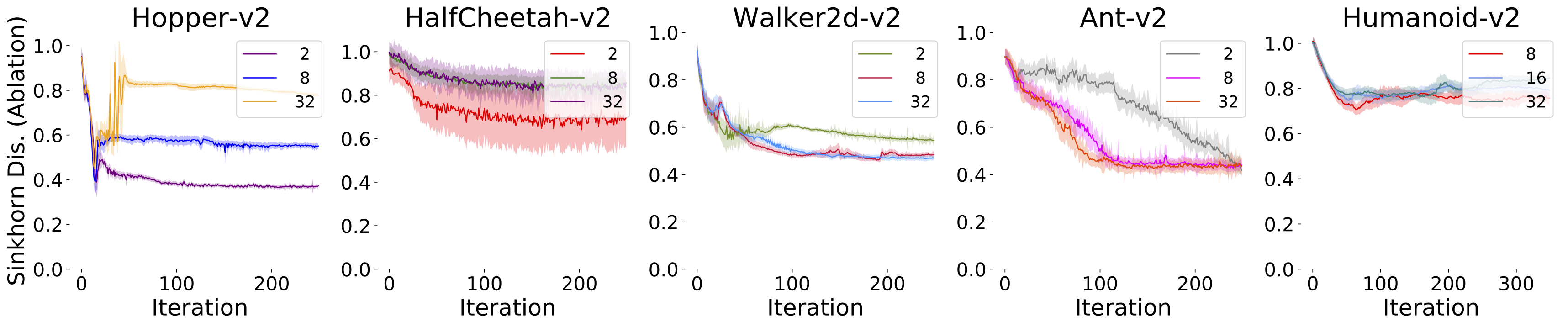}
    \caption{\emph{Ablation Study.} Mean and standard deviation of the Sinkhorn distance during training of SIL for three sets of varying number of trajectories. The critic network update has been replaced with a fixed cosine transport cost defined in Equation (\ref{eq:fixed-cosine}).}
    \label{fig:ablation}
\end{figure}{}
 \begin{table}[t!]
  \begin{center}
    
    \centering
    \resizebox{\textwidth}{!}{
    \begin{tabular}{llcccc}
    \Xhline{1\arrayrulewidth}
      \textbf{Environments}  &\textbf{Metric}  & \textbf{2} & \textbf{8} & \textbf{32} \TBstrut\TBstrut \\
    \Xhline{0.5\arrayrulewidth}
      Hopper-v2              &  Reward &\(264.72\pm1.28\) & \(520.88\pm29.83\) & \(9.44\pm0.31\)\Tstrut \Bstrut\\
                    &   Sinkhorn & \(0.036\pm0.007\) & \(0.552\pm0.008\) & \(0.777\pm0.007\)\Tstrut \Bstrut\\
         HalfCheetah-v2            &  Reward & \(-1643.98\pm198.31\) & \(-844.52\pm267.42\) & \(-1220.92\pm217.86\)\Tstrut \Bstrut\\
                  &   Sinkhorn &\(0.670\pm0.141\) & \(0.841\pm0.035\) & \(0.424\pm0.031\)\Tstrut \Bstrut\\
         Walker2d-v2           &  Reward & \(60.64\pm7.92\) & \(-2.39\pm14.05\) & \(-11.38\pm1.22\)\Tstrut \Bstrut\\
                    &  Sinkhorn & \(0.538\pm0.006\) & \(0.487\pm0.005\) & \(0.466\pm0.009\)\Tstrut \Bstrut\\
    Ant-v2           &  Reward & \(1482.03\pm480.99\) & \(607.87\pm87.09\) & \(114.22\pm123.24\)\Tstrut \Bstrut\\
                    &  Sinkhorn & \(0.398\pm0.090\) & \(0.419\pm0.025\) & \(0.424\pm0.031\)\Tstrut \Bstrut\\
    
    \Xhline{0.5\arrayrulewidth}
          &&&&&  \TBstrut\TBstrut \\
    \Xhline{1\arrayrulewidth}
    \TBstrut\TBstrut& & \textbf{8} & \textbf{16} & \textbf{32} \TBstrut\TBstrut \\
\Xhline{0.5\arrayrulewidth}
      Humanoid-v2           &  Reward & \(447.87\pm31.26\) & \(505.47\pm67.62\) & \(335.48\pm65.14\)\Tstrut \Bstrut\\
                    &  Sinkhorn & \(0.760\pm0.011\) & \(0.789\pm0.013\) & \(0.835\pm0.012\)  \TBstrut\TBstrut \\
    \Xhline{1\arrayrulewidth}
 \end{tabular}}
  \end{center}
      \caption{Mean and standard deviation of the reward and Sinkhorn metric performance after re-training SIL with a \emph{fixed} cosine transport cost defined in Equation (\ref{eq:fixed-cosine}). \label{tab:ablation}}
\vspace{-10mm}
\end{table}
 Figure \ref{fig:ablation} depicts the evolution of the Sinkhorn distance between occupancy measures during training of SIL, after replacing the adversarial objective of the critic network with a fixed transport cost. While the training process is more stable, it fails to achieve good performance in terms both of the Sinkhorn distance metric (Figures~\ref{fig:sil-subsampled-training} and~\ref{fig:ablation}) and reward metric (see Table~\ref{tab:ablation}). The result suggests that the training objective of the critic network has been a crucial part of the proposed algorithm in providing sufficiently strong signals to the learner policy to match the expert's state-action distribution.

\section{Conclusion}\label{sec:conclusion}
In this work we presented Sinkhorn imitation learning (SIL), a solution to optimal transport based imitation learning, by formulating the problem of matching an expert's state-action distribution as minimization of their Sinkhorn distance. We utilized an adversarially trained critic that maps the state-action observations to an adversarially learned feature space. The use of the critic provides a discriminative signal to the learner policy to facilitate the imitation of an expert demonstrator's behavior.  Experiments on 5 MuJoCo environments demonstrate that SIL exhibits competitive performance compared to the baselines.

The Sinkhorn imitation learning framework can be extended in several directions to address current limitations which we aim to study in future work. Currently, SIL's formulation makes it compatible with only on-policy RL methods as computing the Sinkhorn distance necessitates complete trajectories. While SIL is efficient compared to other on-policy adversarial IL benchmarks, it still requires more environment interactions to learn compared to off-policy adversarial IL methods. Hence, it is an interesting future direction to extend SIL to be compatible with off-policy RL algorithms, in line with previous work \cite{dadashi2020,dac-gail,il-off-policy-dist-matching,reddy2020SQIL} to yield a method that both inherits the theoretical benefits of OT while being sample efficient. Additionally, performance of SIL was reported with a fixed critic network structure in all studied experiments. Hence, it is unclear what is the effect of the network architecture in guiding imitation learning.  It will be of practical significance to investigate the impact of different critic network architectures on training stability and computational efficiency, as well as its relationship to the dimension of state-action space. Another interesting research area is to extend the current framework to incorporate the temporal dependence of the trajectory in the construction of the optimal transport coupling and subsequently the reward proxy. We anticipate that this will be a promising direction for improving the sample efficiency and generalization performance of the optimal transport-based adversarial imitation learning framework.

\bibliography{refsbib}
\bibliographystyle{splncs04}

\end{document}